\documentclass[letterpaper, 10 pt, conference]{ieeeconf}  
\usepackage[utf8]{inputenc}
\usepackage[T1]{fontenc}

\IEEEoverridecommandlockouts                              

\usepackage{graphicx} 
\usepackage{times} 
\usepackage{microtype} 
\usepackage{amsmath} 
\usepackage{amssymb}  
\usepackage{cite}
\usepackage{url}
\usepackage{booktabs}
\usepackage{placeins}
\usepackage[table]{xcolor}
\usepackage[hidelinks]{hyperref}
\newcommand{\authoremail}[1]{\mbox{\href{mailto:#1}{#1}}}
\definecolor{resultblue}{HTML}{176B87}
\definecolor{resulttint}{HTML}{EDF5F8}
\definecolor{resultgray}{HTML}{666E77}
\newcommand{\resultgroup}[2]{\rowcolor{resulttint}\multicolumn{#1}{@{}l}{\strut\textcolor{resultblue}{\bfseries #2}}\\}
\newcommand{\resultna}{\textcolor{resultgray}{N/A}}

\newcommand{\resultpm}[2]{\ensuremath{#1}\,{\textcolor{resultgray}{\ensuremath{\pm #2}}}}
\newcommand{\resultnote}[1]{\par\vspace{3pt}{\footnotesize\noindent #1\par}}

\title{\LARGE \bf
Spatial Memory by Agents, for Agents
}

\author{Qianwei Wang$^{2}$, Ce Zhang$^{3}$, Zhanpeng Luo$^{4}$,
Pengpeng Su$^{5}$, Shuo Yang$^{5}$, and Hao Xu$^{1,*}$%
\thanks{$^{*}$Corresponding author: Hao Xu.}%
\thanks{\raggedright $^{1}$Hao Xu is with Nanjing University
(e-mail: \authoremail{xuhao3e8@nju.edu.cn}).}%
\thanks{\raggedright $^{2}$Qianwei Wang is with Princeton University
(e-mail: \authoremail{qw3601@princeton.edu}).}%
\thanks{\raggedright $^{3}$Ce Zhang is with Carnegie Mellon University
(e-mail: \authoremail{cezhang@cs.cmu.edu}).}%
\thanks{\raggedright $^{4}$Zhanpeng Luo is with Johns Hopkins University
(e-mail: \authoremail{zluo45@jh.edu}).}%
\thanks{\raggedright $^{5}$Pengpeng Su and Shuo Yang are with Mondo Robotics
(e-mail: \authoremail{pengpeng.su@mondorobotics.com};
\authoremail{shuo.yang@mondorobotics.com}).}%
}

\hypersetup{
    pdftitle={Spatial Memory by Agents, for Agents},
    pdfauthor={Qianwei Wang, Ce Zhang, Zhanpeng Luo, Pengpeng Su, Shuo Yang, Hao Xu},
    pdfkeywords={spatial memory, coding agents, embodied intelligence, robotics}
}

\begin{document}

\maketitle
\thispagestyle{empty}
\pagestyle{empty}

\begin{abstract}
Spatial memory enables robots to reuse past observations for localization,
navigation, and question answering.
Designing such memory involves jointly deciding what to retain and how to
retrieve it for diverse semantic, metric, and temporal needs.
We investigate whether coding agents can automate this design process through iterative implementation and evaluation.
Under a fixed development budget, a Designer Agent writes and revises code for memory construction and retrieval using feedback from a fixed Query Agent. 
The resulting implementations are frozen before building memories from held-out observation histories, without access to downstream questions, and evaluated across five complementary task tracks using the same Query Agent.
Across three independent design searches, Codex / GPT-6 Astra averages
44.23 on the equally weighted five-track score, versus 29.32 for ReMEmbR,
the strongest evaluated baseline supporting all five tracks, a 50.9\%
relative improvement. Qualitative analysis relates performance differences
to evidence retention and retrieval, while identifying grounding and
evidence-use failures. 
These results demonstrate the potential of coding agents to automate spatial memory design, while revealing task-dependent trade-offs in performance and computational cost.
Project page: \url{https://ba-fa-3344.github.io/}.
\end{abstract}

\section{Introduction}
\label{sec:introduction}

Spatial memory enables robots to use past observations to answer questions unknown during exploration.
After traversing an office, a robot may be asked where a chair is, which cabinet is beside the desk, what the room looks like, or when it last saw an object.

These questions require complementary spatial, semantic, and temporal evidence. Deciding what to retain and how to organize it is difficult: a memory supporting one query type may omit information essential to another.

Existing systems organize observations differently. ReMEmbR~\cite{Anwar2024ReMEmbR} stores captions with poses and timestamps, ConceptGraphs~\cite{Gu2024ConceptGraphs} builds object-centric graphs, DAAAM~\cite{Gorlo2025DAAAM} maintains temporal object/place histories, and FARM~\cite{He2026FARM} retrieves evidence through spatial relations.
Perception models, association strategies, embeddings, and retrieval tools
create a large, coupled design space: useful retrieval requires both retaining
evidence and making it accessible to downstream agents. Repeated implementation,
evaluation, and revision make manual development costly.

\begin{figure}[!t]
    \centering
    \includegraphics[width=\columnwidth]{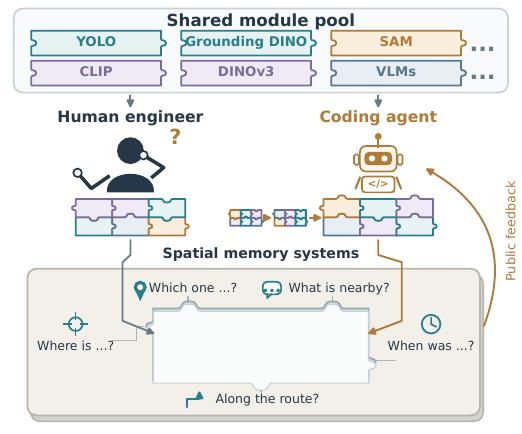}
    \caption{Spatial memory by agents, for agents.
    A shared component pool supports different implementations (puzzles)
    for the same question requirements (board). Coding agents revise their
    designs using public feedback. The human--agent contrast is conceptual.}
    \label{fig:overview}
\end{figure}

Coding agents offer a way to explore this space by implementing candidates,
running experiments, and revising code in response to failures.
Recent work demonstrates these capabilities in agent architectures and ML
pipelines~\cite{Hu2024ADAS,Nam2025MLESTAR}, and in robotics through
trace-guided program repair and real-world policy improvement
~\cite{Lu2026ASPIRE,Xiao2026ENPIRE}. This motivates our question
(Fig.~\ref{fig:overview}): \emph{given reusable components, development
feedback, and a fixed budget, how well can a coding agent design spatial
memory for another agent across diverse tasks?}

\begin{figure*}[!t]
    \centering
    \includegraphics[width=\textwidth]{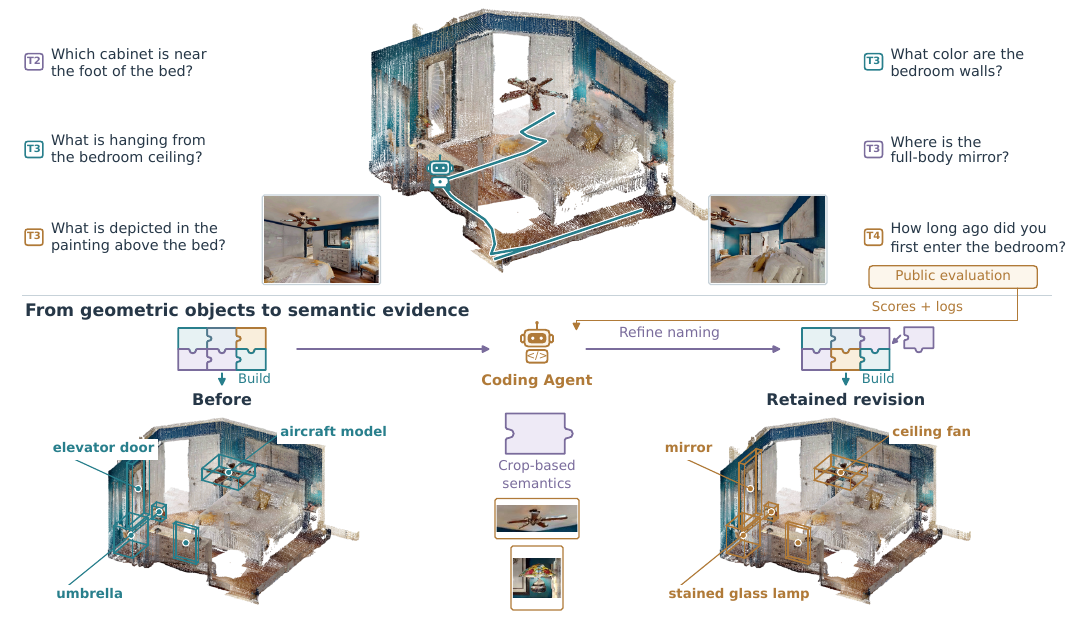}
    \caption{An agent-discovered revision on a public HM3D scene.
    Shared observations and public questions guide a crop-based naming
    revision: \emph{aircraft model} becomes \emph{ceiling fan}, while object
    geometry stays unchanged. Bottom panels show actual memory artifacts,
    not answer predictions; puzzle assemblies denote implementations.}
    \label{fig:design-evolution}
\end{figure*}

Our workflow separates \emph{design}, \emph{build}, and \emph{query}, with distinct Designer and Query Agent roles.
The Designer iteratively develops memory construction and retrieval using public-task feedback.
The selected implementation is frozen before building memories from held-out histories without downstream questions.
A fixed Query Agent then answers through each system's retrieval tools, enabling comparison under a common downstream agent.
Figure~\ref{fig:design-evolution} illustrates how a design revision changes
the semantic evidence stored in memory.

Building on NaVQA and OC-NaVQA~\cite{Anwar2024ReMEmbR,Gorlo2025DAAAM}, our evaluation covers category-level and referring-expression localization, spatial and temporal QA, and navigation-history QA. Across three independent searches, our strongest Designer, Codex / GPT-6 Astra, averages 44.23 Unified on held-out histories versus 29.32 for ReMEmbR~\cite{Anwar2024ReMEmbR}, the strongest evaluated full-track baseline. We analyze implementations, development traces, and resource costs to relate design choices to performance and computational trade-offs.

Our contributions are:
\begin{itemize}
    \item A coding-agent workflow that jointly designs spatial memory construction and retrieval through budgeted, iterative implementation and evaluation.
    \item A five-track spatial memory evaluation suite with separate development and test histories and a fixed Query Agent for controlled cross-task comparisons.
    \item An empirical demonstration of improved average cross-task performance over evaluated full-track baselines, with analyses of evidence retention, retrieval interfaces, and performance--resource trade-offs.
\end{itemize}

\section{Related Work}
\label{sec:related_work}

\subsection{Spatial Memory Construction and Retrieval}

Dynamic scene graphs, Kimera, and Hydra organize geometry, objects,
places, and temporal structure~\cite{Rosinol2020DSG,Rosinol2021Kimera,Hughes2022Hydra};
HOV-SG adds open-vocabulary room--floor hierarchies for navigation
~\cite{Werby2024HOVSG}. Other language-queryable representations include
VLMaps' top-down features~\cite{Huang2023VLMaps}, CLIP-Fields' implicit
fields~\cite{Shafiullah2023CLIPFields}, and OpenScene's distilled point
features~\cite{Peng2023OpenScene}. ConceptFusion fuses multimodal features
into geometry~\cite{Jatavallabhula2023ConceptFusion}, whereas ConceptGraphs
associates objects with descriptions and relations~\cite{Gu2024ConceptGraphs}.

For downstream reasoning, ReMEmbR retrieves posed, timestamped captions
~\cite{Anwar2024ReMEmbR}; 3D-Mem selects visual snapshots
~\cite{Yang2024ThreeDMem}; DAAAM exposes temporal object/place histories
~\cite{Gorlo2025DAAAM}; and FARM performs relational retrieval and visual
verification~\cite{He2026FARM}. Embodied VideoAgent combines appearance,
context, geometry, and history~\cite{Fan2025EmbodiedVideoAgent}, while
GraphEQA couples scene graphs with task-relevant images~\cite{Saxena2025GraphEQA}.

Task-driven representations also have precedents: Clio adjusts map
granularity to a supplied task list~\cite{Maggio2024Clio}. We search the
\emph{implementation} during development and freeze it before building on
held-out histories. Similarly, VLMaps generates code to consume a map,
whereas our Designer implements memory construction and access. Our focus
is the selection and coordination of established mechanisms.

\subsection{Functional Evaluation of Spatial Memory}

Semantic MapNet studies semantic maps and their use in counting and
navigation~\cite{Cartillier2021SemanticMapNet}, MultiON tests multi-object
navigation~\cite{Wani2020MultiON}, and EMQA queries spatial and
spatiotemporal information from observed tours~\cite{Datta2022EMQA}.
NaVQA and OC-NaVQA are the closest precedents for our protocol.
ReMEmbR's NaVQA formulation separates memory construction from spatial,
temporal, and descriptive queries~\cite{Anwar2024ReMEmbR}.
OC-NaVQA annotates actual object positions, uses full observation
sequences, and compares memories using the same reasoning LLM
~\cite{Gorlo2025DAAAM}.
OpenEQA adds open-ended history QA~\cite{Majumdar2024OpenEQA}, and
ScanRefer/ScanEnts3D supply grounded language annotations
~\cite{Chen2019ScanRefer,Abdelreheem2024ScanEnts3D}.
We combine these resources with constructed questions into a shared
build--package--query evaluation, retaining native OC-NaVQA reporting.
The suite tests complementary memory capabilities through development
feedback and held-out assessment for coding-agent design.

\subsection{Coding-Agent-Driven System Design}

ADAS searches code-defined agent architectures
~\cite{Hu2024ADAS}; MLE-STAR uses ablation-guided refinement and ensemble
search for ML pipelines~\cite{Nam2025MLESTAR}; and the Darwin G\"odel
Machine modifies a coding agent's implementation~\cite{Zhang2025DGM}.
MLE-bench evaluates budgeted machine-learning engineering through repeated
agent runs~\cite{Chan2024MLEBench}.
In robotics, ASPIRE repairs robot programs using execution traces and
distills reusable skills~\cite{Lu2026ASPIRE}, while ENPIRE develops reset
and verification interfaces with human feedback to support policy
improvement through real-world trials~\cite{Xiao2026ENPIRE}.
Here, coding agents design persistent memory construction and access,
evaluated across histories and tasks.

\section{Spatial Memory Design with Coding Agents}
\label{sec:methodology}

\subsection{Spatial Memory as the Design Target}
Object grounding uses geometry and identity, while temporal recall uses
event histories
~\cite{Gu2024ConceptGraphs,Anwar2024ReMEmbR,Gorlo2025DAAAM}.
These complementary demands motivate jointly designing memory retention
and access. Given a posed RGB-D history $O=\{(I_t,D_t,P_t,\tau_t)\}$,
a system $s=(B_s,T_s,p_s)$ comprises a builder producing
memory $M_s=B_s(O)$, retrieval tools $T_s$, and tool-use instructions $p_s$.
A shared Query Agent answers through these
interfaces: $a=A_{\theta_0}(q;M_s,T_s,p_s)$.

The shared pool offers diverse pretrained modules, including
YOLOE~\cite{Wang2025YOLOE} for detection, FastSAM for
segmentation, OpenCLIP and DINOv3 for visual
features, Qwen3.5-4B for captioning, and Qwen3-VL-Embedding-2B for
multimodal embeddings.
We favor compact modules and a 4B consumer for onboard efficiency;
quantized Qwen3.5-4B supports responsive inference on Jetson Orin
devices~\cite{NVIDIA2026EdgeLLM}.

\subsection{Designer Protocol}
Each scratch search starts from an unimplemented builder with shared task
instructions, metrics, module pool, and public data. The Designer may inspect
public data and prior work and edit the builder, retrieval tools, and tool-use
instructions; the evaluator and Query Agent core remain fixed.

The loop is \emph{implement--evaluate--inspect--keep/revert--revise}.
Separate H100 workers build candidates and run Qwen3.5-4B evaluation,
returning per-track scores, costs, predictions, and query-linked tool traces.
The Designer can inspect public reference answers and memory packages,
then record keep/revert decisions with rationales. Selection uses only
public feedback.

The 24-hour budget covers coding, API waiting, building, and evaluation.
Configurations share task/evaluation interfaces but retain native harness
editing, execution, and context management. Full prompts, tools,
configurations, and execution rules are in the code.

We summarize progress with a resource-adjusted public score:
\begin{equation}
    \mathcal{J}_{\mathrm{report}}(s)=
    \frac{1}{5}\sum_{t=1}^{5}u_t(s)
    -C_{\mathrm{resource}}(s),
    \label{eq:design-objective}
\end{equation}
where $u_t\in[0,1]$ is the public track score. With $[x]_+=\max(0,x)$,
the resource penalty is
\[
 C_{\mathrm{resource}}(s)=0.2[m_s/m_0-1]_+
 +0.5\min\{1,[b_s/b_0-1]_+\}.
\]
Here $m_s$ and $b_s$ average package bytes and build TPF across public
inputs; $m_0=50\times10^6$ bytes and $b_0=0.2$\,s/frame. Both penalties
vanish within budget; the time penalty caps at 0.5. Query latency is not
penalized.

\subsection{Frozen Build--Package--Query Evaluation}
The complete implementation is frozen before held-out evaluation. Packages
are built from histories without questions or references and reused across
tracks. Systems share the Query Agent checkpoint, common instructions,
interaction loop, decoding, tool budget, retries, and query order;
only method-specific tools and their instructions differ.

\section{Multifaceted Evaluation}
\label{sec:evaluation}

\subsection{Coverage, Sources, and Split}
Five tracks probe complementary memory requirements: category coverage
and geometry (T1), instance discrimination (T2), semantic context (T3),
temporal evidence (T4), and navigation-history QA (T5).

\begin{figure}[!t]
    \centering
    \includegraphics[width=\columnwidth]{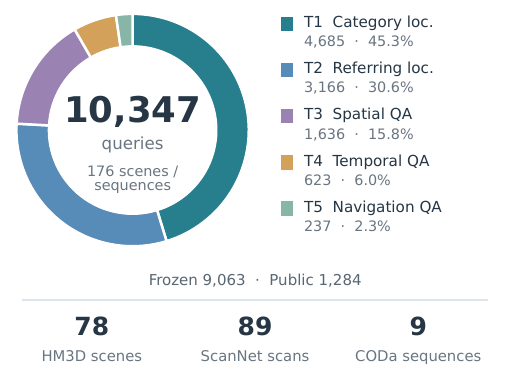}
    \caption{Benchmark composition across unique scenes and sequences.
    Sectors show query shares; Unified weights the five tracks equally.}
    \label{fig:distribution}
\end{figure}

We combine constructed questions with ScanNet and HM3D
inputs~\cite{Dai2017ScanNet,Ramakrishnan2021HM3D}, ScanRefer expressions
and ScanEnts3D annotations~\cite{Chen2019ScanRefer,Abdelreheem2024ScanEnts3D},
OpenEQA~\cite{Majumdar2024OpenEQA}, and OC-NaVQA~\cite{Gorlo2025DAAAM}
on CODa~\cite{Zhang2024CODa}. Figure~\ref{fig:distribution} summarizes
10,347 queries over 176 inputs: 167 spatial inputs and nine sequences.
Public development contains 1,284 queries; frozen evaluation contains
9,063. Complete scene-scan/episode/trajectory IDs are disjoint across
splits, although separate captures may revisit physical regions.
T5 includes 177 official frozen questions and 60 constructed,
human-reviewed public questions on separate trajectories.

\subsection{What Each Score Measures}
Scores lie in $[0,1]$; missing or malformed answers receive zero.

\paragraph{T1: Category-level metric localization}
The system returns instance positions for a category. A prediction at a
3D distance $d$ from a target center receives credit $w(d)=\max(0,1-d/g)$, where
$g=\operatorname{clip}(\tfrac12\text{bbox diagonal},0.5,2.0)$ meters.
Each query specifies a prediction cap $K$.
After maximum-credit one-to-one matching of only the first $K$
predictions, let $G$ and $P$ count targets and predictions,
$R=\min(G,K)$, and $C$ be total credit. The soft F-score is the harmonic
mean of $C/R$ and $C/\max(P,R)$, measuring localization and instance coverage.

\paragraph{T2: Referring-expression localization}
Only the top-ranked position receives $w(d)$ against the referred target,
measuring instance resolution and localization.

\paragraph{T3: Open-ended spatial QA}
A reference-guided Qwen3.5-4B judge assigns a semantic correctness rating
$r\in\{1,\ldots,5\}$, mapped to $(r-1)/4$. It uses a separate conversation from the Query Agent,
but shares its checkpoint.

\paragraph{T4: Temporal QA}
For absolute time error $e$ and observation horizon $H$ in seconds,
$\ell(e,H)=\operatorname{clip}(1-\log(1+e)/\log(1+H),0,1)$ rewards
temporal precision relative to history length. For multiple events, the
$i$-th prediction is paired with the $i$-th reference; scores are averaged
over references, with missing events scored zero and extras ignored.

\paragraph{T5: Navigation QA}
Recorded-history questions use exact yes/no accuracy,
$\operatorname{clip}(1-d_{xy}/(100\,\mathrm{m}),0,1)$ for horizontal
position error, or $\ell(e,H)$ for time/duration converted to seconds.
Original OC-NaVQA errors are also reported separately; this track does
not measure active navigation success.

\subsection{Aggregation and Resource Accounting}
Each track uses a query-micro average; \emph{Unified} equally averages
the five tracks. Unsupported tracks are N/A, with no Unified for partial
systems. Formal results use only frozen queries. Resources are reported
separately: build time per frame, package size, peak RAM/VRAM, and
end-to-end query latency. The shared Query Agent uses temperature zero,
disabled thinking, a 2,048-token output limit, and five tool iterations.


\newcommand{\MainResultsTable}{%
\begin{table*}[!t]\centering
\caption{Five-track results and design cost. Three independent searches per Designer.}\label{tab:main}
\begingroup\footnotesize
\setlength{\tabcolsep}{1.5pt}\renewcommand{\arraystretch}{1.1}
\begin{tabular*}{\textwidth}{@{\extracolsep{\fill}}lrrrrrrrrr@{}}
\toprule
\multicolumn{8}{c}{Frozen performance} & \multicolumn{2}{c}{24-hour design search} \\
\cmidrule(lr){1-8}\cmidrule(l){9-10}
System / Designer &  & T1 $\uparrow$ & T2 $\uparrow$ & T3 $\uparrow$ & T4 $\uparrow$ & T5 $\uparrow$ & Unified $\uparrow$ & Tokens (M) $\downarrow$ & Cost (\$) $\downarrow$ \\
\midrule
LLM-with-Captions &  & \resultpm{0.87}{0.00} & \resultpm{0.56}{0.00} & \resultpm{59.70}{0.06} & \resultpm{19.49}{0.38} & \resultpm{51.27}{0.13} & \resultpm{26.38}{0.10} & \resultna & \resultna \\
ConceptGraphs &  & \resultpm{2.07}{0.02} & \resultpm{6.49}{0.02} & \resultpm{31.98}{0.14} & \resultna & \resultna & \resultna & \resultna & \resultna \\
DAAAM &  & \resultpm{9.48}{0.00} & \resultpm{17.97}{0.06} & \resultpm{40.33}{0.42} & \resultpm{31.38}{0.33} & \resultpm{38.64}{0.51} & \resultpm{27.56}{0.18} & \resultna & \resultna \\
ReMEmbR &  & \resultpm{1.01}{0.00} & \resultpm{0.47}{0.01} & \resultpm{61.64}{0.11} & \resultpm{24.49}{0.17} & \resultpm{58.98}{0.62} & \resultpm{29.32}{0.18} & \resultna & \resultna \\
FARM-4B &  & \resultpm{16.43}{0.01} & \resultpm{25.77}{0.08} & \resultpm{46.18}{0.15} & \resultna & \resultna & \resultna & \resultna & \resultna \\
\midrule
\textcolor{resultblue}{\bfseries \shortstack[l]{Claude /\\Claude Opus 4.8}} & Best & \resultpm{20.37}{0.04} & \resultpm{28.97}{0.12} & \resultpm{54.70}{0.10} & \resultpm{55.43}{0.30} & \resultpm{52.01}{0.29} & \resultpm{42.30}{0.11} & 599 & 452 \\
 & Avg & \resultpm{17.65}{2.38} & \resultpm{24.96}{6.38} & \resultpm{54.05}{2.08} & \resultpm{53.70}{2.00} & \resultpm{51.37}{2.91} & \resultpm{40.35}{2.80} & \resultpm{631}{67} & \resultpm{463}{44} \\
\textcolor{resultblue}{\bfseries \shortstack[l]{Claude /\\Claude Opus 5}} & Best & \resultpm{17.53}{0.08} & \resultpm{24.79}{0.07} & \resultpm{53.45}{0.19} & \resultpm{\mathbf{56.39}}{0.04} & \resultpm{\mathbf{60.12}}{0.30} & \resultpm{42.46}{0.10} & 755 & 561 \\
 & Avg & \resultpm{19.31}{1.91} & \resultpm{29.53}{4.45} & \resultpm{48.39}{6.81} & \resultpm{52.90}{3.97} & \resultpm{53.15}{6.04} & \resultpm{40.66}{1.61} & \resultpm{690}{84} & \resultpm{520}{56} \\
\textcolor{resultblue}{\bfseries \shortstack[l]{Codex /\\GPT-6 Astra}} & Best & \resultpm{\mathbf{22.33}}{0.06} & \resultpm{\mathbf{39.14}}{0.14} & \resultpm{\mathbf{61.66}}{0.28} & \resultpm{52.81}{0.25} & \resultpm{52.78}{0.37} & \resultpm{\mathbf{45.75}}{0.14} & \textbf{302} & 428 \\
 & Avg & \resultpm{20.97}{1.62} & \resultpm{37.55}{1.60} & \resultpm{61.35}{0.96} & \resultpm{49.30}{3.81} & \resultpm{52.00}{1.27} & \resultpm{44.23}{1.54} & \resultpm{287}{14} & \resultpm{406}{20} \\
\textcolor{resultblue}{\bfseries \shortstack[l]{DeepSeek /\\DeepSeek-V4-Pro}} & Best & \resultpm{22.26}{0.03} & \resultpm{30.44}{0.07} & \resultpm{48.38}{0.14} & \resultpm{45.51}{0.31} & \resultpm{53.03}{0.43} & \resultpm{39.92}{0.16} & 564 & \textbf{25} \\
 & Avg & \resultpm{18.50}{3.49} & \resultpm{31.03}{0.76} & \resultpm{47.01}{4.21} & \resultpm{47.24}{2.08} & \resultpm{50.07}{2.61} & \resultpm{38.77}{1.40} & \resultpm{494}{67} & \resultpm{23}{3} \\
\bottomrule
\end{tabular*}\endgroup
\resultnote{Scores $\times100$, mean $\pm$ sample std. Baselines/Best: three query repeats. Avg: three independent search means, each averaging three query repeats; std measures design-search variability. Avg is primary; Best uses highest public Unified within each configuration. Bold excludes Avg. Design tokens/cost: one search for Best, three-search mean $\pm$ std for Avg. N/A: unsupported/not applicable.}
\end{table*}
}

\newcommand{\ResourceTable}{%
\begin{table*}[!t]\centering
\caption{Build and query costs ($\downarrow$).}\label{tab:resources}
\begingroup\footnotesize
\setlength{\tabcolsep}{1.5pt}\renewcommand{\arraystretch}{1.1}
\begin{tabular*}{\textwidth}{@{\extracolsep{\fill}}lrrrrrr@{}}
\toprule
System &  & \shortstack{Build TPF\\s/frame} & \shortstack{Package\\MiB} & \shortstack{Peak RAM\\GiB} & \shortstack{Peak VRAM\\GiB} & \shortstack{Query\\s} \\
\midrule
LLM-with-Captions &  & \resultpm{0.047}{0.027} & 0.14 & 6.6 & 27.1 & \resultpm{9.56}{0.11} \\
ConceptGraphs &  & \resultpm{0.078}{0.060} & 0.23 & 6.6 & 11.4 & \resultpm{6.17}{0.37} \\
DAAAM &  & \resultpm{0.667}{0.387} & 67.90 & 17.3 & 22.8 & \resultpm{11.27}{0.06} \\
ReMEmbR &  & \resultpm{0.049}{0.028} & 2.07 & 8.4 & 33.3 & \resultpm{9.96}{0.46} \\
FARM-4B &  & \resultpm{0.425}{0.101} & 67.77 & 15.6 & 47.0 & \resultpm{5.98}{0.08} \\
\midrule
Claude / Claude Opus 4.8 & Best & \resultpm{0.171}{0.084} & 3.98 & 10.8 & 31.6 & \resultpm{19.19}{0.18} \\
 & Avg & \resultpm{0.166}{0.104} & 6.19 & 11.4 & 34.3 & \resultpm{14.87}{3.38} \\
Claude / Claude Opus 5 & Best & \resultpm{0.113}{0.068} & 3.37 & 11.4 & 36.5 & \resultpm{12.43}{0.37} \\
 & Avg & \resultpm{0.143}{0.079} & 4.31 & 10.3 & 27.1 & \resultpm{14.74}{2.72} \\
Codex / GPT-6 Astra & Best & \resultpm{0.180}{0.062} & 15.19 & 10.4 & 33.1 & \resultpm{26.14}{1.14} \\
 & Avg & \resultpm{0.253}{0.100} & 23.94 & 11.7 & 32.7 & \resultpm{25.93}{1.41} \\
DeepSeek / DeepSeek-V4-Pro & Best & \resultpm{0.149}{0.076} & 7.51 & 5.0 & 6.3 & \resultpm{13.91}{0.74} \\
 & Avg & \resultpm{0.347}{0.449} & 10.67 & 10.0 & 23.6 & \resultpm{14.89}{1.77} \\
\bottomrule
\end{tabular*}\endgroup
\resultnote{TPF: per-input mean $\pm$ sample std, counting all input frames and excluding model-pool startup. Package and per-input peak RAM/VRAM: means only. Query: mean $\pm$ sample std across repeat means. Avg pools three systems; these stds describe input/repeat dispersion. Partial systems query only T1--T3. Hardware: H100 for build and Qwen query.}
\end{table*}
}

\newcommand{\TransferTable}{%
\begin{table*}[!t]\centering
\caption{Seeded initialization: per-track initial and final performance.}\label{tab:transfer}
\begingroup\footnotesize
\setlength{\tabcolsep}{1.5pt}\renewcommand{\arraystretch}{1.1}
\begin{tabular*}{\textwidth}{@{\extracolsep{\fill}}lrrrrrrr@{}}
\toprule
Seed & Run & \shortstack{T1\\initial $\to$ final} & \shortstack{T2\\initial $\to$ final} & \shortstack{T3\\initial $\to$ final} & \shortstack{T4\\initial $\to$ final} & \shortstack{T5\\initial $\to$ final} & \shortstack{Unified\\initial $\to$ final} \\
\midrule
ReMEmbR & 1 & 1.01 $\to$ 14.51 & 0.47 $\to$ 31.23 & 61.64 $\to$ 61.04 & 24.49 $\to$ 48.52 & 58.98 $\to$ 65.81 & 29.32 $\to$ 44.22 \\
ReMEmbR & 2 & 1.01 $\to$ 17.05 & 0.47 $\to$ 26.62 & 61.64 $\to$ 50.65 & 24.49 $\to$ 54.40 & 58.98 $\to$ 51.18 & 29.32 $\to$ 39.98 \\
FARM-4B & 1 & 16.43 $\to$ 15.42 & 25.77 $\to$ 27.90 & 46.18 $\to$ 32.76 & \resultna $\to$ 46.16 & \resultna $\to$ 47.56 & \resultna $\to$ 33.96 \\
FARM-4B & 2 & 16.43 $\to$ 21.70 & 25.77 $\to$ 30.65 & 46.18 $\to$ 51.17 & \resultna $\to$ 48.16 & \resultna $\to$ 46.34 & \resultna $\to$ 39.60 \\
\bottomrule
\end{tabular*}\endgroup
\resultnote{Scores $\times100$, means of three frozen query repeats. Initial scores reuse the baselines in Table~\ref{tab:main}; final scores follow two searches per seed with Claude / Claude Opus 5. FARM-4B initially supports T1--T3; all final systems support T1--T5.}
\end{table*}
}

\newcommand{\DeploymentTable}{%
\begin{table}[!t]\centering
\caption{Simulation robot deployment: 40 house tasks.}\label{tab:deployment}
\begingroup\footnotesize
\setlength{\tabcolsep}{1.5pt}\renewcommand{\arraystretch}{1.1}
\begin{tabular*}{\columnwidth}{@{\extracolsep{\fill}}lrr@{}}
\toprule
System & \shortstack{Navigation SR\\20 tasks, \% $\uparrow$} & \shortstack{Memory-QA Acc.\\20 tasks, \% $\uparrow$} \\
\midrule
No Memory & \resultpm{3.3}{2.9} & \resultpm{10.0}{5.0} \\
Keyframe RAG & \resultpm{6.7}{7.6} & \resultpm{31.7}{2.9} \\
ReMEmbR & \resultpm{10.0}{5.0} & \resultpm{\mathbf{36.7}}{2.9} \\
\shortstack[l]{Codex /\\GPT-6 Astra Best} & \resultpm{\mathbf{58.3}}{5.8} & \resultpm{33.3}{5.8} \\
\bottomrule
\end{tabular*}\endgroup
\resultnote{Mean $\pm$ sample std across inference seeds 42, 43, and 44; 20 tasks per category and seed. Controller, task starts, and prebuilt memories are fixed.}
\end{table}
}

\newcommand{\OCNavTable}{%
\begin{table}[!t]\centering
\caption{OC-NaVQA: 177 official frozen queries.}\label{tab:ocnavqa}
\begingroup\footnotesize
\setlength{\tabcolsep}{1.5pt}\renewcommand{\arraystretch}{1.1}
\begin{tabular*}{\columnwidth}{@{\extracolsep{\fill}}lcccc@{}}
\toprule
System & \shortstack{\strut Binary\\\strut \% $\uparrow$} & \shortstack{\strut Pos.\\\strut m $\downarrow$} & \shortstack{\strut Time err.\\\strut min $\downarrow$} & \shortstack{\strut Query\\\strut s $\downarrow$} \\
\midrule
\resultgroup{5}{Qwen3.5-4B}
\shortstack[l]{LLM-with-\\Captions} & \resultpm{64.18}{0.00} & \resultpm{63.44}{1.48} & \resultpm{8.34}{0.19} & \resultpm{7.56}{0.15} \\
DAAAM & \resultpm{38.31}{0.86} & \resultpm{57.26}{3.77} & \resultpm{5.11}{1.08} & \resultpm{48.83}{0.93} \\
ReMEmbR & \resultpm{70.15}{1.49} & \resultpm{\mathbf{50.97}}{0.51} & \resultpm{\mathbf{4.97}}{0.08} & \resultpm{\mathbf{7.00}}{0.61} \\
\shortstack[l]{Codex /\\GPT-6 Astra Best} & \resultpm{59.70}{0.00} & \resultpm{56.25}{2.44} & \resultpm{5.59}{0.41} & \resultpm{19.94}{2.51} \\
\shortstack[l]{Claude / Claude\\Opus 5 Best} & \resultpm{\mathbf{78.61}}{0.86} & \resultpm{51.55}{0.00} & \resultpm{5.82}{0.06} & \resultpm{9.79}{0.83} \\
\resultgroup{5}{GPT-5-mini}
\shortstack[l]{LLM-with-\\Captions} & \resultpm{72.64}{3.76} & \resultpm{40.79}{5.80} & \resultpm{3.93}{0.51} & \resultpm{17.09}{0.07} \\
DAAAM & \resultpm{65.17}{4.80} & \resultpm{48.33}{3.26} & \resultpm{3.51}{0.15} & \resultpm{46.98}{1.95} \\
ReMEmbR & \resultpm{72.14}{3.11} & \resultpm{\mathbf{24.84}}{2.88} & \resultpm{\mathbf{1.75}}{0.15} & \resultpm{19.26}{0.60} \\
\shortstack[l]{Codex /\\GPT-6 Astra Best} & \resultpm{77.61}{4.48} & \resultpm{44.32}{4.07} & \resultpm{2.36}{0.25} & \resultpm{20.84}{1.46} \\
\shortstack[l]{Claude / Claude\\Opus 5 Best} & \resultpm{\mathbf{79.60}}{0.86} & \resultpm{55.31}{3.15} & \resultpm{4.38}{0.26} & \resultpm{\mathbf{13.00}}{1.03} \\
\bottomrule
\end{tabular*}\endgroup
\resultnote{Mean $\pm$ sample std over three 177-query repeats; bold: best per answerer. Query time includes completion attempts. DAAAM alone uses its native five-tool, serial 15-call protocol here for GPT-5-mini; the main evaluation uses five tool iterations. Qwen: H100; GPT: local tools and remote API.}
\end{table}
}

\newcommand{\SearchFigure}{%
\begin{figure}[!t]\centering
\includegraphics[width=\columnwidth]{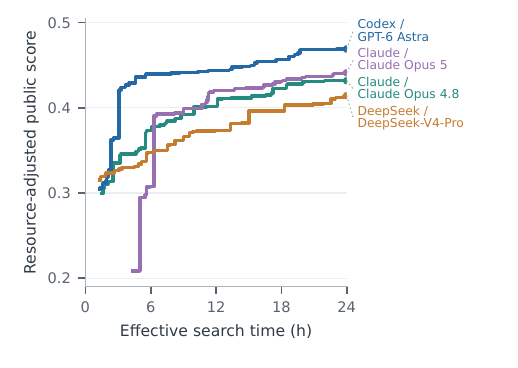}
\caption{Public Unified minus resource penalty: mean best-retained progress across three searches using eight H100 GPUs.}\label{fig:search}
\end{figure}
}

\MainResultsTable

\section{Experiments and Analysis}
\label{sec:experiments}

\subsection{Setup and Reporting}
Each of four harness--model configurations performs three independent
24-hour searches. Eight H100 GPUs per search parallelize builds and
Qwen3.5-4B evaluations for faster feedback.
Claude uses Max effort; Codex and DeepSeek use High.
Each system builds once per input and serves three query repeats.
Scores are scaled by 100. Avg, our primary comparison, reports mean
$\pm$ sample std across three search means, measuring design-search
variability (Table~\ref{tab:main}). Each mean averages three query repeats.
Best is the run with the highest public Unified within each configuration.
All checkpoint/run selection uses public feedback; frozen evaluation follows.

To reduce model-related variation, baselines share the Qwen3.5-4B Query
Agent and use common build components where possible.
Our ConceptGraphs implementation uses YOLO-World,
SAM, and OpenCLIP from the shared pool. ReMEmbR retains caption--pose--time
memory, using the same Qwen3.5-4B captioner and Qwen3-VL-Embedding-2B
embedder available to the Designers. For the same reason, FARM's
Qwen3.5-9B captioner and query parser are replaced with Qwen3.5-4B;
we call this configuration FARM-4B. DAAAM retains its own mapping stack,
including FastSAM, DAM, visual/text encoders, and tracking, and serves the
shared Query Agent. LLM-with-Captions provides a caption-memory control.
ConceptGraphs and FARM-4B are evaluated on their supported T1--T3.

\ResourceTable
\subsection{Effectiveness and Cost}
We first compare cross-task memory utility and deployment cost under the
shared Query Agent.
\emph{Broader memory utility across tasks.}
Codex / GPT-6 Astra averages 44.23 Unified across three searches versus
29.32 for ReMEmbR, the strongest full-track baseline:
a 14.91-point (50.9\% relative) gain (Table~\ref{tab:main}). The other
configuration averages range from 38.77 to 40.66; all 12 systems exceed
ReMEmbR on Unified. Astra's average exceeds DAAAM
on all five tracks and FARM-4B on its three supported tracks. On T2,
it reaches 37.55 versus 17.97 and 25.77, respectively. Its gains over
ReMEmbR concentrate in localization and temporal QA; T3 is nearly tied,
and ReMEmbR remains stronger on T5. Opus 5 has higher average T4/T5 scores
than Astra, illustrating complementary task profiles. The Astra run selected
by public Unified achieves 45.75 frozen Unified; Avg remains the primary
across-search comparison.

\emph{Compact memories with different query costs.}
Agent-designed systems also use compact storage (Table~\ref{tab:resources}).
Mean package sizes span 4.31--23.94\,MiB across configurations,
compared with 67.90 for DAAAM and 67.77 for FARM-4B. Mean build TPF
is also lower: 0.143--0.347\,s/frame versus 0.667/0.425.
ReMEmbR remains smaller at 2.07\,MiB.
Astra averages 23.94\,MiB and 25.93\,s/query, compared with Opus 5's
4.31\,MiB and 14.74\,s/query, despite lower average T4/T5 scores.
The workflow finds storage-efficient systems, but query-time reasoning
adds a performance--cost trade-off.

\subsection{Designed Systems and Performance Differences}
We inspect all 12 implementations and public packages alongside the
baselines to interpret their task profiles. This \emph{qualitative}
comparison uses the systems selected by public Unified as concrete examples
of spatial evidence, temporal granularity, and retrieval choices.

\emph{Preserving the right spatial quantity.}
ReMEmbR stores observer positions alongside captions, whereas Codex / GPT-6 Astra Best
retains object geometry, linked views, and alternative crops. Localization
accounts arithmetically for 12.00 of their 16.43-point Unified gap:
an observation location need not be the described object's location.
Its target appearance, full-query context, and conditional target--anchor
parsing support instance discrimination. FARM-4B shares geometric and
relational retrieval ingredients; the important design choices concern
instance retention, context representation, and ranking together.

\emph{Representing events at the required granularity.}
DAAAM provides object histories, region visits, and trajectory queries.
Claude / Claude Opus 5 Best makes further task-facing distinctions among
sightings, encounters, passing, area visits, and level arrivals in
revisit-aware places. Its T4 score of 56.39 versus DAAAM's 31.38 is
consistent with the value of matching event semantics to the question.
Codex / GPT-6 Astra Best retains temporal evidence too, but is stronger on
T1--T3 and weaker on T4/T5. These complementary profiles reflect the
different evidence needed for object discrimination and event interpretation.

\emph{Coordinating complexity with the consumer.}
Claude / Claude Opus 4.8 Best combines geometric fusion, visual/lexical ranking, caption
reranking, and neighboring-object evidence. DeepSeek / DeepSeek-V4-Pro Best takes a more
compact route: an instance/crop store, temporal region layer, and three
tools. Its seven-tool Run~2 scores 0.75 points lower on Unified while
requiring almost three times the build TPF, although it improves T2--T4.
More components offer capacity but only help when their evidence remains
usable within the shared interaction budget.

\emph{Remaining challenges in grounding and evidence use.}
Reliable instance grounding remains a central challenge: even the best
T1/T2 scores are 22.33/39.14. These tasks couple coverage, instance
discrimination, and position accuracy. Temporal QA adds a semantic
difficulty: a precise time can describe the wrong event.
In a public DeepSeek / DeepSeek-V4-Pro Run~3 candidate, the Query
Agent uses a 119.25\,s visibility span to answer a bathroom-dwell question
whose reference is 7.1\,s. Finally, retained evidence needs an effective
access path. Context overflows and invalid tool arguments in selected
DeepSeek systems cause failed frozen queries; the development traces below
also show useful geometry going unused. These challenges span both memory
representation and its use by the agent.

\TransferTable
\OCNavTable
\DeploymentTable
\SearchFigure
\subsection{Search Process and Evidence Use}
Figure~\ref{fig:search} tracks progress with eight H100 GPUs per search. Between-system variation exceeds fixed-system
query variation, highlighting the value of independent searches.

\emph{Readout changes can unlock existing evidence.}
For a public living-room duration question (19.7\,s ground truth), Codex /
GPT-6 Astra Run~2 initially nests visit spans beneath scene views; the
Query Agent answers with the 113.3\,s trajectory horizon. Promoting the
spans to primary records changes the answer to 13\,s; stationary-camera
support improves it to 15.333\,s, with the same build and tool. Readout
improves evidence use, although the answer remains imperfect.

Conversely, a revision requires an extra inspection call for coordinates.
The Query Agent makes none across 1,284 public queries; T2 falls from
35.49 to zero and the revision is reverted. The geometry exists but its
access path fails. Figure~\ref{fig:design-evolution} illustrates the
complementary refinement of stored semantic evidence.

\subsection{Seeded Initialization}
Given the scratch results, we next ask whether an established memory
system provides a better starting point for broader task coverage.
Two Claude / Claude Opus 5 searches start from each of ReMEmbR and FARM-4B,
using the scratch budget and public-only development protocol
(Table~\ref{tab:transfer}). Both ReMEmbR outcomes improve T1, T2, and T4,
addressing its weak localization and temporal scores. Run~1 largely
preserves T3 and improves T5, reaching 44.22 Unified versus 42.46 for the
matched scratch Best; Run~2 loses T3/T5 performance. The two outcomes
illustrate the challenge of extending a memory while retaining its strengths.

Both FARM-4B outcomes add T4/T5 support. On its original T1--T3, however,
Run~1 improves only T2, whereas Run~2 improves all three. Both remain below
the matched scratch Avg of 40.66. Seeding thus offers useful starting capabilities,
with mixed outcomes as the Designer adapts them to broader demands.

\subsection{OC-NaVQA and Answerer Compatibility}
To connect the five-track results with earlier spatial-memory studies,
Table~\ref{tab:ocnavqa} reports the official 177-question subset within T5
using native OC-NaVQA metrics. DAAAM's comparison uses
GPT-5-mini~\cite{Gorlo2025DAAAM}, motivating a second evaluation of the same
packages with this Query Agent alongside our shared Qwen3.5-4B setup.
This also lets us examine how memory utility changes with the consumer.

Claude / Claude Opus 5 Best leads binary accuracy with both answerers
(78.61/79.60\%). Codex / GPT-6 Astra Best rises from 59.70\% with Qwen3.5-4B
to 77.61\% with GPT-5-mini,
versus 72.14\% for ReMEmbR and 65.17\% for DAAAM. It also lowers DAAAM's
position/time errors from 48.33\,m/3.51\,min to 44.32\,m/2.36\,min, although
ReMEmbR remains best on both errors (24.84\,m/1.75\,min). The native-metric
view thus preserves the complementary strengths seen in the main results:
agent-designed memories are competitive, while accurate metric and temporal
answers remain sensitive to the memory--consumer pairing. The table also
reports query costs for each configuration.

\begin{figure}[!t]
    \centering
    \includegraphics[width=\columnwidth]{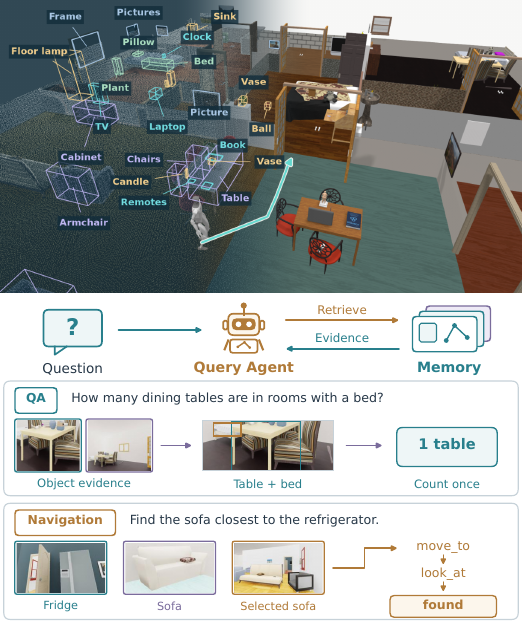}
    \caption{A Query Agent reasons with memory from Codex / GPT-6 Astra Best.
    Below the illustrative house overview, recorded examples show contextual
    QA and target-directed navigation.}
    \label{fig:simulation-deployment}
\end{figure}

\subsection{Simulation Robot Deployment}
The recorded-history results motivate a further question: can the designed
memory also guide a robot's actions? We test navigation and QA in a
ProcTHOR house~\cite{Deitke2022ProcTHOR}
within Isaac Sim (Fig.~\ref{fig:simulation-deployment}). No Memory, Keyframe
RAG, ReMEmbR, and Codex / GPT-6 Astra Best share a Qwen3.5-4B controller
and task starts; memories are prebuilt from one robot survey and read-only.
No Memory uses only current observations and task-local interaction;
Keyframe RAG retrieves sampled survey images by embedding similarity,
together with their poses and timestamps.

The 40 tasks comprise 20 navigation goals and 20 QA questions covering
counting, distance, and support relations.
The controller receives live observations, pose, a known collision map,
and each system's native retrieval tools. Navigation succeeds when the
robot declares the target found within 1\,m with depth-confirmed
visibility; QA uses exact integer or choice matching.
Table~\ref{tab:deployment} reports three inference-seed repeats.

\emph{Turning memory into navigation targets.}
The agent-designed memory achieves 58.3\% navigation success versus
10.0\% for ReMEmbR. Its object coordinates and crops provide actionable
targets: the successful sofa episode in Fig.~\ref{fig:simulation-deployment}
retrieves sofa/refrigerator candidates, then moves to and faces the selected
sofa. ReMEmbR's caption poses instead locate observations, not objects.
Target confirmation remains difficult: in 20 of 25 failed episodes,
the robot declares success prematurely; five are near the correct target
but fail depth-based visibility. Improving this final stage requires
connecting retrieved evidence with the robot's current view and approach.

\emph{Context and coverage in memory QA.}
QA is 33.3\% versus ReMEmbR's 36.7\%: only two additional correct answers
for ReMEmbR across 60 repeated trials. Distance comparisons favor ReMEmbR
(9/18 versus 7/18); captions can preserve useful co-occurrence without
object coordinates. Both systems struggle with complete, unique counts.
For the table-count question in Fig.~\ref{fig:simulation-deployment},
the agent-designed system answers two instead of one in two repeats;
the illustrated repeat inspects context and counts once. Bounded retrieval
and instance/context ambiguity can impede enumeration: whole-house QA
requires consistent evidence across objects and rooms, beyond a useful
navigation target.

\FloatBarrier
\section{Discussion}
\label{sec:discussion}

\emph{Conclusion.}
Spatial memory benefits from coordinated retention, retrieval, and use.
Coding-agent design improves average cross-task utility over established
baselines, producing compact memories with complementary localization,
temporal reasoning, and navigation strengths.

\emph{Limitations.}
Complete-system evaluation leaves individual mechanisms less understood;
controlled ablations would deepen our qualitative analysis.
The onboard-oriented pool and 4B consumer leave stronger cloud-hosted
reasoners and hybrid designs largely unexplored. OC-NaVQA's answerer-dependent
results motivate these comparisons and independent T3 judges.

\emph{Future Work.}
How should robotics benchmarks evaluate agent-adapted systems? Pairing task
metrics with public feedback, held-out environments, explicit budgets, and
agent-designed baselines could test new mechanisms, component coordination,
and efficiency under changing demands.
\section*{Acknowledgments}
We used coding agents to help write code and create figures.
\bibliographystyle{IEEEtran}
\bibliography{references}

\end{document}